\documentclass[12pt]{article}
\usepackage[utf8]{inputenc}
\usepackage[a4paper, margin=1in]{geometry}
\usepackage{graphicx}
\usepackage{float}
\usepackage{multirow}
\usepackage{booktabs}
\usepackage{amsmath}
\usepackage{hyperref}
\usepackage{caption}

\title{Evaluating Large Language Models for Simulated Ophthalmic Decision-Making Using Retinal Fundus Photographs}

\author{
Cindy Tabuse$^{1}$ \and
David Restrepo$^{2,3}$ \and
Carolina Gracitelli$^{1}$ \and
Fernando Korn Malerbi$^{1}$ \and
Caio Regatieri$^{1}$ \and
Luis Filipe Nakayama$^{1,3}$
}

\date{
\small
$^1$Ophthalmology Department, São Paulo Federal University, São Paulo – SP, Brazil\\
$^2$Université Paris-Saclay, CentraleSupélec, Mathematics and Computer Science Laboratory\\
$^3$Laboratory for Computational Physiology, Massachusetts Institute of Technology, Cambridge, MA, USA
}
\begin{document}

\maketitle

\begin{abstract}
\textbf{Background:} Large language models (LLMs) can simulate clinical reasoning based on natural language prompts, but their utility in ophthalmology is largely unexplored. This study evaluated GPT-4’s ability to interpret structured textual descriptions of retinal fundus photographs and simulate clinical decisions for diabetic retinopathy (DR) and glaucoma screening, including the impact of adding real or synthetic clinical metadata.

\textbf{Methods:} A retrospective diagnostic validation study was conducted using 300 annotated fundus images. GPT-4 received structured prompts describing each image, with or without real or synthetic patient metadata. The model was asked to assign an ICDR severity score, recommend DR referral, and estimate the cup-to-disc ratio for glaucoma referral. Model performance was evaluated using accuracy, macro and weighted F1 scores, and Cohen’s kappa. McNemar’s test and change rate analysis assessed the statistical and practical impact of metadata inclusion.

\textbf{Results:} GPT-4 demonstrated moderate performance for ICDR severity classification, with an accuracy of 67.5\%, macro F1 score of 0.33, weighted F1 score of 0.67, and Cohen’s kappa of 0.25. The correct identification of normal cases largely drove performance. When reframed as a binary referral task for DR, performance improved (accuracy 82.3\%, F1 score 0.54, kappa 0.44). In contrast, performance for glaucoma referral was poor across all conditions (accuracy of 78\%, F1 score <0.04, kappa <0.03). The inclusion of metadata, either real or synthetic, had no statistically significant effect on model outputs (McNemar’s $p > 0.05$), and predictions remained largely consistent across conditions, with change rates under 7\% for ICDR and under 3\% for DR referral.

\textbf{Discussion:} GPT-4 can simulate basic ophthalmic decision-making using structured prompts but lacks precision in complex tasks like glaucoma screening or fine-grained DR grading. Its predictions rely primarily on textual image descriptions rather than metadata. While not clinically reliable, LLMs may assist in educational, documentation, or annotation tasks in ophthalmology.
\end{abstract}

\section{Introduction}

Diabetic retinopathy (DR) remains one of the leading causes of preventable blindness worldwide, particularly among individuals with longstanding diabetes \cite{sun2022idf, teo2021global}. Early detection and timely referral are critical to avoid irreversible visual impairment \cite{teo2021global}. However, the global burden of diabetes continues to rise, placing strain on eye care services, especially in regions where access to retinal specialists and high-cost imaging infrastructure is limited. Traditional remote DR screening relies on the acquisition of retinal fundus photographs, which are evaluated by trained graders or automated systems for signs of disease \cite{nakayama2023review}. While advances in deep learning have led to the development of highly accurate image-based diagnostic models, these systems require training on large annotated image datasets, and their deployment typically depends on specialized infrastructure.

Beyond diabetic retinopathy, glaucoma is another leading cause of irreversible blindness worldwide, characterized by progressive optic neuropathy often associated with elevated intraocular pressure \cite{weinreb2014pathophysiology}. Glaucoma frequently remains asymptomatic until advanced stages, underscoring the importance of early detection, especially in populations already undergoing retinal screening \cite{tan2023evaluating}. Incorporating glaucoma screening into diabetic eye evaluations has been proposed as a cost-effective strategy to expand access to eye care \cite{crowston2004effect}. Simple surrogate markers, such as the assessment of the optic disc cup-to-disc ratio (CDR) from fundus photographs, can serve as a preliminary screening tool to identify patients at risk for glaucoma who may benefit from comprehensive ophthalmologic evaluation. However, this task remains subjective, with substantial inter-grader variability, even among specialists.

In parallel with progress in image-based artificial intelligence (AI), large language models (LLMs) have emerged as general-purpose AI systems capable of performing a wide range of tasks based solely on natural language inputs \cite{brown2020language}. Unlike computer vision models that directly process pixel data, LLMs operate by interpreting textual descriptions, contextual information, and structured prompts to generate outputs that simulate reasoning and decision-making processes. These models have demonstrated strong performance across diverse applications, including medical question-answering, clinical report summarization, and guideline interpretation \cite{kung2023usmle, gobira2023chatgpt}. However, the use of LLMs in ophthalmology remains largely unexplored \cite{antaki2023evaluating}.

Despite this potential, there are significant gaps in understanding how LLMs perform in tasks that mimic clinical reasoning based on retinal findings. It is also unknown how the inclusion of demographic and clinical metadata influences LLM decision-making when evaluating ophthalmic conditions. Metadata such as age, sex, hypertension, obesity, smoking status, and insulin use are well-established modifiers of disease prevalence and severity in diabetic retinopathy and glaucoma, and their incorporation may impact model outputs.

This study aims to evaluate the performance of an LLM in interpreting retinal fundus photographs through structured natural language prompts that simulate the clinical assessment of diabetic and glaucomatous patients. Specifically, the model was tasked with assigning an International Clinical Diabetic Retinopathy (ICDR) severity score, making referral recommendations for diabetic retinopathy based on established clinical thresholds, estimating the cup-to-disc ratio, and determining whether a referral for glaucoma evaluation is warranted. To investigate the influence of metadata, the model was tested under three conditions: using image descriptions alone, with real patient metadata, and with synthetically generated metadata. Importantly, the goal of this study is not to assess the LLM as a clinical diagnostic tool, but rather to explore its capabilities and limitations in performing structured ophthalmic tasks based on language prompts. This research provides early insights into how LLMs may be integrated into ophthalmology for non-diagnostic applications such as clinical documentation, education, and scalable data annotation.

\section{Methods}

\subsection{Study Design and Objective}
\begin{figure}[H]
  \centering
  \includegraphics[width=0.9\textwidth]{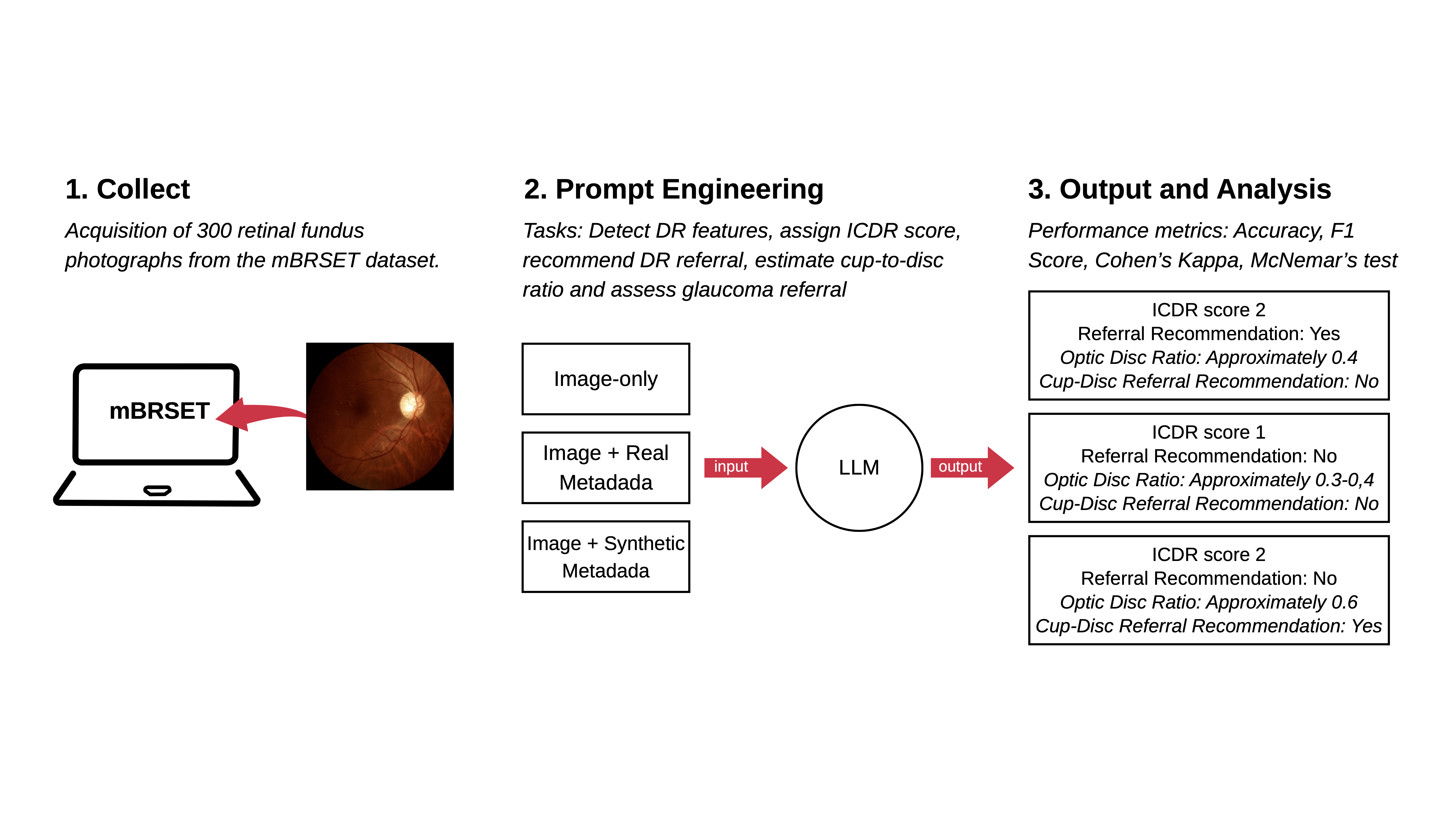}
  \caption{Experiments and evaluation framework.}
  \label{fig:framework}
\end{figure}

This was a retrospective diagnostic validation study designed to assess the performance of LLMs in the automated classification of DR severity, referral recommendations, and cup-to-disc ratio evaluation based on retinal fundus photographs. The study compared the model’s performance when provided with (i) image-only prompts, (ii) real clinical metadata, and (iii) synthetic metadata to evaluate the impact of structured demographic and clinical information on classification outcomes (Figure~\ref{fig:framework}).

\subsection{Dataset}
The source dataset was the Brazilian Ophthalmological Dataset (mBRSET), which contains annotated retinal fundus photographs with labels for diabetic retinopathy severity, referral recommendations, and optic nerve evaluations~\cite{nakayama2023review, wu2025portable}. From this dataset, a subset of 300 images was randomly selected for this study. Each image was paired with corresponding demographic and clinical data, including age, sex, insulin use, systemic hypertension, obesity, and smoking status.

\subsection{Prompt Engineering}
A structured prompt was designed to simulate the decision-making process of an ophthalmologist evaluating a retinal exam from a diabetic patient. The prompt instructed the model to (Appendix):
\begin{enumerate}
    \item Identify diabetic retinopathy features, including microaneurysms, hemorrhages, exudates, macular edema, and laser scars indicative of treated proliferative disease.
    \item Assign an International Clinical Diabetic Retinopathy (ICDR) severity score based on the identified findings. The presence of panretinal photocoagulation scars was considered sufficient to classify as proliferative diabetic retinopathy (ICDR level 4).
    \item Recommend referral if the ICDR score was $\geq 2$ and/or if macular edema was present.
    \item Estimate the optic disc cup-to-disc ratio and recommend glaucoma referral if the ratio exceeds 0.6.
\end{enumerate}

The output included the ICDR score, a findings summary, referral recommendation for diabetic retinopathy, measured cup-to-disc ratio, and referral recommendation for glaucoma.

\subsection{Model Architecture and Access}
All prompts were submitted to GPT-4, a state-of-the-art large language model developed by OpenAI~\cite{openai2023gpt4}. The model was accessed via the ChatGPT platform, using the "gpt-4" engine under default system settings, with temperature and sampling parameters left unchanged. GPT-4 is a transformer-based autoregressive model trained on a broad range of internet and domain-specific data, but it does not have direct access to image inputs or external databases. Therefore, all image content was conveyed through structured textual descriptions. The model does not retain memory between prompts, and each prompt was processed independently to avoid contextual contamination across cases. This approach ensures replicability of the experiment under controlled prompt design and fixed system parameters.

\subsection{Metadata Conditions}
Three experimental conditions were applied to each image:
\begin{itemize}
  \item \textbf{Image-Only Prompt:} The prompt contained only the image description without any additional clinical information.
  \item \textbf{Real Metadata Prompt:} The prompt incorporated real patient demographics and clinical data extracted from the dataset.
  \item \textbf{Synthetic Metadata Prompt:} A synthetic metadata generator created plausible but randomized demographic and clinical profiles for each image, maintaining realistic distributions. For age, 20\% of records were assigned as “$\geq$90,” and the remaining were random integers between 20 and 89. Binary clinical variables (sex, insulin use, hypertension, obesity, and smoking) were randomly assigned.
\end{itemize}

In the metadata conditions, the prompt included a narrative description combining the image analysis request with patient demographic context (Appendix).

\subsection{Model Evaluation}
The LLM was tasked with performing three classification tasks:
\begin{itemize}
  \item ICDR Multiclass Classification (severity levels 0--4)~\cite{wilkinson2003proposed}.
  \item Referral for Diabetic Retinopathy (binary classification: referable vs. non-referable).
  \item Referral for Glaucoma (binary classification based on cup-to-disc ratio)~\cite{crowston2004effect, tan2023evaluating}.
\end{itemize}

\begin{figure}[H]
  \centering
  \includegraphics[width=0.9\textwidth]{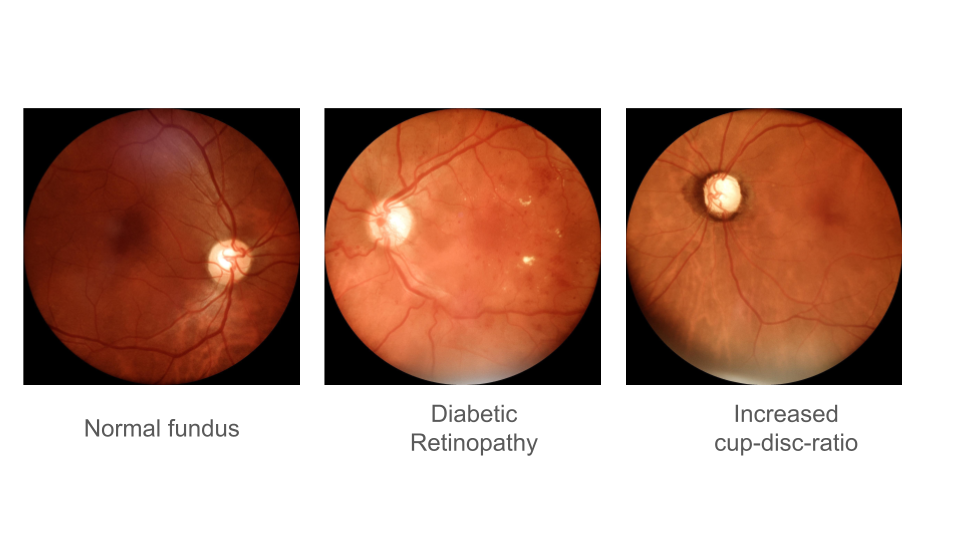}
  \caption{Imaging examples of normal fundus, diabetic retinopathy and increased cup-disc-ratio exams.}
  \label{fig:examples}
\end{figure}

Outputs from each experimental condition were compared to the reference labels provided in the dataset (Figure~\ref{fig:examples}).

\subsection{Statistical Analysis}
Statistical analysis included the calculation of accuracy, macro F1 score, weighted F1 score, and Cohen’s kappa coefficient to assess overall model performance. For the ICDR multiclass classification task, per-class precision, recall, and F1 scores were also computed to evaluate class-specific performance.

To evaluate the contribution of metadata to model performance, pairwise comparisons were conducted between image-only predictions and metadata-based predictions. Additionally, to assess the impact of altering metadata, predictions based on real metadata were compared to those generated using synthetic metadata. These comparisons were performed using McNemar’s test, a non-parametric method for paired categorical data, which tests for statistically significant differences in the distribution of prediction errors between models.

In addition to McNemar’s test, the degree of concordance between prediction pairs was assessed using two complementary approaches. First, the change rate was calculated as the proportion of cases where the predicted ICDR score or referral decision differed between the two conditions being compared. Second, Cohen’s kappa coefficient was computed to quantify the agreement between paired predictions beyond chance. Cohen’s kappa coefficient ranges from -1 to 1, with higher values indicating stronger agreement beyond chance. Values above 0.60 are generally considered substantial, and those above 0.80 reflect near-perfect agreement. This analysis was applied to both the ICDR multiclass grading and the binary referral decisions for diabetic retinopathy and glaucoma.

All statistical analyses were performed using Python (version 3.9), employing the \texttt{scikit-learn} library for performance metrics and kappa statistics, and the \texttt{statsmodels} library for McNemar’s test.

\section{Results}

The ground truth distribution of the dataset reflected a predominance of normal cases for DR and a higher proportion of referable cases for glaucoma compared to DR. For the ICDR classification, the majority of images were labeled as no diabetic retinopathy (class 0) [198 cases, 74.72\%], followed by moderate non-proliferative DR (class 2) [27 cases, 10.19\%], mild non-proliferative DR (class 1) [22 cases, 8.30\%], proliferative DR (class 4) [15 cases, 5.66\%], and severe non-proliferative DR (class 3) [3 cases, 1.13\%]. For the binary classification of referral for DR, 83.02\% were labeled as non-referable [220 cases] and 16.98\% as referable [45 cases]. In contrast, referral for glaucoma showed a higher proportion of referable cases, with 77.74\% classified as non-referable [206 cases] and 22.26\% as referable [59 cases].

\subsection{ICDR Multiclass Classification}

The performance of the models for ICDR grading demonstrated limited agreement with the gold standard. The image-based model achieved a kappa of 0.250 and an accuracy of 0.675. Macro F1 was 0.330, and weighted F1 was 0.672. Performance was primarily driven by the correct classification of normal cases (class 0; F1 = 0.82), while performance on referable and advanced stages was poor, with F1 scores of 0.00 for classes 1 and 3.

The metadata-based model yielded comparable results (kappa = 0.246, accuracy = 0.679, macro F1 = 0.372, weighted F1 = 0.675). McNemar’s test comparing metadata to image-only predictions resulted in a p-value of 1.00, indicating no statistically significant difference between the two approaches.

Similarly, the use of synthetic metadata resulted in slightly lower performance (kappa = 0.225, accuracy = 0.653, macro F1 = 0.358, weighted F1 = 0.659), but McNemar’s test comparing real vs. synthetic metadata (p = 0.0923) also showed no significant difference, suggesting that metadata alone—whether real or synthetic—does not meaningfully improve or degrade ICDR classification performance relative to image-based predictions.

Pairwise analysis of ICDR predictions revealed that the use of real metadata compared to image-only predictions led to a change in ICDR grading in 6.79\% of cases (kappa = 0.614), while changing from image-only to synthetic metadata resulted in a 7.17\% change (kappa = 0.594). The comparison between real and synthetic metadata showed a change rate of 6.04\% (kappa = 0.658), indicating moderate agreement but non-negligible variability in predicted ICDR scores across conditions.

\subsection{Referral for Diabetic Retinopathy (Binary Classification)}

Reframing the task to binary referral (referable vs. non-referable) substantially improved model performance compared to the multiclass ICDR task. The image-based model achieved a kappa of 0.436, an accuracy of 0.823, and an F1 score of 0.544. McNemar’s test comparing image and metadata predictions yielded a p-value of 0.7266, indicating no significant difference between them.

The metadata-based model achieved similar results (kappa = 0.433, accuracy = 0.830, F1 = 0.536), as did the synthetic metadata model (kappa = 0.419, accuracy = 0.819, F1 = 0.529). McNemar’s test comparing real vs. synthetic metadata produced a p-value of 0.3750, again showing no significant difference. These results indicate that both image features and metadata—whether real or synthetic—perform comparably for DR referral detection.

Change rate analysis demonstrated high consistency in referral predictions for diabetic retinopathy. The referral decision changed in only 3.02\% of cases when comparing image-based to real metadata predictions (kappa = 0.908) and in 1.89\% of cases when comparing image-based to synthetic metadata (kappa = 0.944). The agreement between real and synthetic metadata predictions was similarly high, with a change rate of 1.89\% (kappa = 0.942), indicating that the addition or modification of metadata had minimal impact on the referral decision for DR.

\subsection{Referral for Glaucoma (Binary Classification)}

All models exhibited poor performance for glaucoma referral. The image-based model produced a kappa of 0.026, an accuracy of 0.781, and an F1 score of 0.033. Despite the low performance, McNemar’s test between image and metadata predictions resulted in a p-value of 1.0000, indicating no statistically significant difference between the two methods.

The metadata-based model similarly achieved poor performance (kappa = 0.000, accuracy = 0.777, F1 = 0.000), as did the synthetic metadata model (kappa = 0.004, accuracy = 0.770, F1 = 0.032). McNemar’s test comparing real vs. synthetic metadata resulted in a p-value of 0.6250, further confirming no significant difference. These findings underscore that neither image nor metadata-based models are reliable for glaucoma referral within this dataset.

Nevertheless, change rate analysis revealed that the referral decision for glaucoma was generally stable between image and real metadata predictions, changing in only 0.38\% of cases (kappa = 0.000). However, when comparing image-based predictions to those using synthetic metadata, the referral decision changed in 1.89\% of cases (kappa = -0.006), and in 1.51\% of cases when comparing real to synthetic metadata (kappa = 0.000). These extremely low kappa values reflect the poor baseline performance of all models for glaucoma referral rather than meaningful disagreement (Table 1).

\begin{table}[H]
\centering
\caption{Comparison metrics across tasks, including McNemar's p-values for pairwise comparisons.}
\resizebox{\textwidth}{!}{%
\begin{tabular}{llccccc}
\toprule
\textbf{Task} & \textbf{Condition} & \textbf{Kappa} & \textbf{Accuracy} & \textbf{Macro F1} & \textbf{Weighted F1} & \textbf{McNemar p-value} \\
\midrule
\multirow{3}{*}{ICDR (Multiclass)} 
 & Image-Only         & 0.250 & 0.675 & 0.330 & 0.672 & 1.0000 (vs Metadata) \\
 & Real Metadata      & 0.246 & 0.679 & 0.372 & 0.675 & 0.0923 (vs Synthetic) \\
 & Synthetic Metadata & 0.225 & 0.653 & 0.358 & 0.659 & – \\
\midrule
\multirow{3}{*}{DR Referral (Binary)} 
 & Image-Only         & 0.436 & 0.823 & 0.544 & 0.831 & 0.7266 (vs Metadata) \\
 & Real Metadata      & 0.433 & 0.830 & 0.536 & 0.835 & 0.3750 (vs Synthetic) \\
 & Synthetic Metadata & 0.419 & 0.819 & 0.529 & 0.827 & – \\
\midrule
\multirow{3}{*}{Glaucoma Referral (Binary)} 
 & Image-Only         & 0.026 & 0.781 & 0.033 & 0.689 & 1.0000 (vs Metadata) \\
 & Real Metadata      & 0.000 & 0.777 & 0.000 & 0.680 & 0.6250 (vs Synthetic) \\
 & Synthetic Metadata & 0.004 & 0.770 & 0.032 & 0.683 & – \\
\bottomrule
\end{tabular}%
}
\end{table}

\label{tab:comparison_metrics}

\section{Discussion}

This study provides an early evaluation of the potential and limitations of large language models (LLMs), specifically GPT-4, in performing structured ophthalmic tasks based on natural language prompts derived from retinal fundus photographs. The findings suggest that while GPT-4 is capable of simulating certain aspects of clinical reasoning when provided with detailed textual descriptions, its performance varies depending on the complexity of the task and the availability or quality of clinical metadata.

For the classification of DR severity using the ICDR scale, the model demonstrated limited agreement with the reference standard \cite{gulshan2016development, ting2017development}. The image-based prompt achieved moderate accuracy in distinguishing no retinopathy from more advanced stages but showed poor ability to differentiate between intermediate severity levels, such as mild and moderate non-proliferative DR. This is consistent with challenges seen in image-based deep learning models, where subtle clinical signs frequently lead to misclassification at lower severity thresholds \cite{ting2017development}. Notably, performance improved when the task was reframed as a binary classification for referral versus non-referral, suggesting that LLMs may be better suited for simplified decision-support tasks rather than detailed grading of DR severity.

The evaluation of optic nerve characteristics, specifically the estimation of CDR and glaucoma referral recommendations, revealed substantial limitations. Model performance was consistently poor across all configurations. This likely reflects the inherent difficulty of translating subjective, fine-grained spatial assessments of the optic nerve into textual descriptions that an LLM can interpret reliably. It is worth noting that even among glaucoma specialists, there is often considerable disagreement when evaluating the optic nerve, due to the disease’s highly variable presentation and the broad range of normal anatomical variations \cite{garway1998interobserver, cheung2018diagnostic}. These results underscore that tasks requiring detailed spatial interpretation, such as optic nerve evaluation, remain a challenge for language-based models in their current form.

A key finding from this study is the limited impact of clinical metadata on model performance. Pairwise comparisons between image-based and metadata-based predictions showed no statistically significant differences across tasks. Furthermore, comparisons between real and synthetic metadata similarly demonstrated no significant differences. This was further supported by change rate analysis, which showed that the referral decision for diabetic retinopathy changed in only 1.9\% to 3.0\% of cases when transitioning between image, real metadata, and synthetic metadata conditions, with Cohen’s kappa values consistently above 0.9, reflecting near-perfect agreement. For ICDR grading, although the task was inherently more complex, the change rates between methods remained modest, indicating moderate consistency despite the known challenges of multiclass DR severity classification. In contrast, the consistently poor performance and extremely low kappa values observed in the glaucoma referral task underscore that the models failed to capture relevant decision patterns regardless of input condition, reflecting limitations not only in the model but likely also in the task formulation itself.

This indicates that the model’s outputs are primarily driven by the descriptive content of the image-based prompts rather than the structured clinical context provided by metadata. These results suggest that GPT-4, in this context, does not function as a true multimodal model. Instead, it relies predominantly on general medical knowledge encoded during pretraining, with limited capacity to dynamically integrate structured patient-specific metadata into its predictions. This highlights important limitations in the current application of LLMs for personalized clinical reasoning and raises broader concerns regarding the risks of overestimating their capacity for contextual understanding, with implications for fairness, bias, and interpretability in medical AI.

It is critical to emphasize that the use of GPT-4 in this study is not intended for direct clinical prediction or diagnosis. Unlike image-based deep learning systems, which are specifically trained on pixel-level data and validated for clinical use, LLMs operate exclusively on textual inputs and outputs. Their architecture is inherently designed for language understanding, not for image interpretation. Nonetheless, the ability of LLMs to simulate clinical reasoning from structured descriptions opens up potential applications in ophthalmology for non-diagnostic tasks, including standardized report generation, assisting with dataset annotation for AI development, or serving as educational tools to support clinical reasoning \cite{liu2023evaluating}.

These findings should be interpreted in light of several limitations. First, the input relied on textual descriptions of retinal images, which inherently reduces the granularity of clinical detail compared to direct image analysis. Second, the prompts were based on idealized and standardized descriptions; in real-world practice, variability in how findings are documented could further influence model performance. Third, this study evaluated a single LLM at one point in time using a static interface. Performance may differ with future model iterations or alternative LLM architectures. Additionally, the analysis was limited to a subset of the same dataset (mBRSET) without external validation, restricting the assessment of generalizability across populations, devices, or documentation styles. Finally, the use of synthetic metadata, while methodologically useful, may introduce implausible clinical scenarios, potentially confounding the model and underestimating the value of authentic clinical context.

Despite these limitations, this study demonstrates that LLMs can engage with structured ophthalmic tasks to a meaningful extent when provided with carefully designed prompts. While they are not a substitute for image-based diagnostic models and are not intended for clinical decision-making, LLMs may serve a complementary role in ophthalmology by supporting research workflows, scaling dataset annotation, generating standardized reports, or assisting with clinical documentation. Future work should explore the development of truly multimodal AI systems that combine the visual processing capabilities of image-based models with the reasoning abilities of LLMs, offering the potential for more comprehensive and flexible decision-support tools in ophthalmology.

\clearpage
\appendix
\section*{Appendix: Example of Real and Synthetic Metadata Input}
\addcontentsline{toc}{section}{Appendix: Example of Real and Synthetic Metadata Input}

\subsection*{Real Metadata Prompt}

\begin{verbatim}
Please review the retinal exam of a male diabetic patient, with 58 years, uses insulin, 
has systemic hypertension, is not obese, does not smoke, using the clinical details provided above. 
Use the criteria below to guide your analysis:

Retinal Assessment
Identify signs of diabetic retinopathy, including microaneurysms, hemorrhages, 
exudates (both cotton wool spots and hard exudates), and retinal edema.
Specifically note any presence of macular edema.
If panretinal (panphotocoagulation) laser scars are present, consider this a sign 
of treated proliferative diabetic retinopathy.

Severity Evaluation (ICDR Score)
Assign an ICDR score based on the abnormalities detected.
Note: The presence of panphotocoagulation scars qualifies the eye as having 
proliferative diabetic retinopathy (ICDR level 4), regardless of other active signs.
Provide a concise explanation of the key features that influenced your scoring decision.

Referral Recommendation
Recommend referral for further specialist evaluation if the ICDR score is greater 
than or equal to 2 and/or if macular edema is present.
Clearly state whether a referral is indicated.

Optic Disc Ratio
Measure the Cup-to-Disc Ratio by identifying the optic disc and the cup within it, 
and assess the ratio of the cup's diameter to the disc's diameter.
A cup-to-disc ratio greater than 0.6 may indicate potential glaucoma.

Cup-Disc Referral Recommendation
If the cup-to-disc ratio is elevated (greater than 0.6), recommend referral for a 
comprehensive glaucoma evaluation.
Clearly state whether a referral for glaucoma is indicated.

Output Format:
ICDR Score: [Your Score]
Findings: [Summary of abnormalities and features observed]
Referral Recommendation: [Yes/No – with a brief explanation]
Optic Disc Ratio: [Measured ratio – with comments]
Cup-Disc Referral Recommendation: [Yes/No – with a brief explanation]
\end{verbatim}

\subsection*{Synthetic Metadata Prompt}

\begin{verbatim}
Please review the retinal exam of a female diabetic patient, with 22 years, 
does not use insulin, does not have systemic hypertension, is obese, smokes, 
using the clinical details provided above. Use the criteria below to guide your analysis:

Retinal Assessment
Identify signs of diabetic retinopathy, including microaneurysms, hemorrhages, 
exudates (both cotton wool spots and hard exudates), and retinal edema.
Specifically note any presence of macular edema.
If panretinal (panphotocoagulation) laser scars are present, consider this a sign 
of treated proliferative diabetic retinopathy.

Severity Evaluation (ICDR Score)
Assign an ICDR score based on the abnormalities detected.
Note: The presence of panphotocoagulation scars qualifies the eye as having 
proliferative diabetic retinopathy (ICDR level 4), regardless of other active signs.
Provide a concise explanation of the key features that influenced your scoring decision.

Referral Recommendation
Recommend referral for further specialist evaluation if the ICDR score is greater 
than or equal to 2 and/or if macular edema is present.
Clearly state whether a referral is indicated.

Optic Disc Ratio
Measure the Cup-to-Disc Ratio by identifying the optic disc and the cup within it, 
and assess the ratio of the cup's diameter to the disc's diameter.
A cup-to-disc ratio greater than 0.6 may indicate potential glaucoma.

Cup-Disc Referral Recommendation
If the cup-to-disc ratio is elevated (greater than 0.6), recommend referral for a 
comprehensive glaucoma evaluation.
Clearly state whether a referral for glaucoma is indicated.

Output Format:
ICDR Score: [Your Score]
Findings: [Summary of abnormalities and features observed]
Referral Recommendation: [Yes/No – with a brief explanation]
Optic Disc Ratio: [Measured ratio – with comments]
Cup-Disc Referral Recommendation: [Yes/No – with a brief explanation]
\end{verbatim}


\begin{thebibliography}{99}

\bibitem{sun2022idf}
Sun H, Saeedi P, Karuranga S, et al.
IDF Diabetes Atlas: Global, regional and country-level diabetes prevalence estimates for 2021 and projections for 2045.
\textit{Diabetes Res Clin Pract.} 2022;183:109119.

\bibitem{teo2021global}
Teo ZL, Tham YC, Yu M, et al.
Global Prevalence of Diabetic Retinopathy and Projection of Burden through 2045: Systematic Review and Meta-analysis.
\textit{Ophthalmology.} 2021;128(11):1580-1591.

\bibitem{nakayama2023review}
Nakayama LF, Ribeiro LZ, Novaes F, et al.
Artificial intelligence for telemedicine diabetic retinopathy screening: a review.
\textit{Ann Med.} 2023;55(2):2258149.

\bibitem{weinreb2014pathophysiology}
Weinreb RN, Aung T, Medeiros FA.
The pathophysiology and treatment of glaucoma: a review.
\textit{JAMA.} 2014;311(18):1901-1911.

\bibitem{tan2023evaluating}
Tan RE, Teo KYC, Husain R, et al.
Evaluating the outcome of screening for glaucoma using colour fundus photography-based referral criteria in a teleophthalmology screening programme for diabetic retinopathy.
\textit{Br J Ophthalmol.} 2023.

\bibitem{crowston2004effect}
Crowston JG, Hopley CR, Healey PR, et al.
The effect of optic disc diameter on vertical cup to disc ratio percentiles in a population based cohort: the Blue Mountains Eye Study.
\textit{Br J Ophthalmol.} 2004;88(6):766-770.

\bibitem{brown2020language}
Brown T, Mann B, Ryder N, et al.
Language models are few-shot learners.
\textit{Adv Neur Inf Proc Sys.} 2020;33:1877–1901.

\bibitem{kung2023usmle}
Kung TH, Cheatham M, Medenilla A, et al.
Performance of ChatGPT on USMLE: Potential for AI-assisted medical education using large language models.
\textit{PLoS Digit Health.} 2023;2(2):e0000198.

\bibitem{gobira2023chatgpt}
Gobira M, Nakayama LF, Moreira R, et al.
Performance of ChatGPT-4 in answering questions from the Brazilian National Examination for Medical Degree Revalidation.
\textit{Rev Assoc Med Bras.} 2023;69(10).

\bibitem{antaki2023evaluating}
Antaki F, Touma S, Milad D, et al.
Evaluating the performance of ChatGPT in ophthalmology: An analysis of its successes and shortcomings.
\textit{Ophthalmol Sci.} 2023;3(4):100324.

\bibitem{gulshan2016development}
Gulshan V, Peng L, Coram M, et al.
Development and validation of a deep learning algorithm for detection of diabetic retinopathy in retinal fundus photographs.
\textit{JAMA.} 2016;316(22):2402–2410.

\bibitem{ting2017development}
Ting DSW, Cheung CY, Lim G, et al.
Development and validation of a deep learning system for diabetic retinopathy and related eye diseases using retinal images from multiethnic populations with diabetes.
\textit{JAMA.} 2017;318(22):2211–2223.

\bibitem{garway1998interobserver}
Garway-Heath DF, Ruben S, Viswanathan A, Hitchings RA.
Interobserver agreement for the clinical assessment of optic discs.
\textit{Br J Ophthalmol.} 1998;82(6):644–647.

\bibitem{cheung2018diagnostic}
Cheung CY, Rulli E, Schulze A, et al.
Diagnostic accuracy of optic nerve head imaging for glaucoma: a systematic review and meta-analysis.
\textit{Lancet Digit Health.} 2018;1(4):e172–e182.

\bibitem{liu2023evaluating}
Liu X, Zhang T, Chen M, et al.
Evaluating Large Language Models for Simulated Ophthalmic Decision-Making.
\textit{Ophthalmol Sci.} 2023.

\bibitem{openai2023gpt4}
OpenAI.
GPT-4 Technical Report.
\textit{arXiv preprint arXiv:2303.08774.} 2023.

\bibitem{nakayama2024mbrset}
Nakayama LF, Malerbi FK, Regatieri CV, et al.
mBRSET: The Brazilian Ophthalmological Dataset for Diabetic Retinopathy and Multimodal Retinal AI.
\textit{In preparation.} 2024.

\bibitem{stolz2023llms}
Stolz D, Keestra S, et al.
Large Language Models in Medicine: Current Limitations and Future Potential.
\textit{npj Digit Med.} 2023.

\bibitem{wu2025portable}
Wu C, Restrepo D, Nakayama LF, et al.
A portable retina fundus photos dataset for clinical, demographic, and diabetic retinopathy prediction.
\textit{Sci Data.} 2025;12(1):323.

\bibitem{wilkinson2003proposed}
Wilkinson CP, Ferris FL, Klein RE, et al.
Proposed international clinical diabetic retinopathy and diabetic macular edema disease severity scales.
\textit{Ophthalmology.} 2003;110(9):1677–1682.

\end{thebibliography}
\end{document}